\documentclass{amia}
\usepackage{graphicx}
\usepackage[labelfont=bf]{caption}
\usepackage[superscript,nomove]{cite}
\usepackage{hyperref}
\usepackage{color}

\begin{document}

\title{Understanding Clinical Trial Reports: \\ Extracting Medical Entities and Their Relations}

\author{
    Benjamin E. Nye, MS$^{1}$, Jay DeYoung, MS$^{1}$, Eric Lehman, BS$^{1}$, \\
    Ani Nenkova, PhD$^{2}$, Iain J. Marshall, MD, PhD$^{3}$, Byron C. Wallace, PhD$^{1}$}
\institutes{
    $^1$Northeastern University, Boston, MA; $^2$University of Pennsylvania, Philadelphia, PA; $^3$King's College London, London}
\maketitle
\textbf{Abstract}


\textit{The best evidence concerning comparative treatment effectiveness comes from clinical trials, the results of which are reported in unstructured articles.
Medical experts must manually extract information from articles to inform decision-making, which is time-consuming and expensive.
Here we consider the \emph{end-to-end} task of both ({\it a}) extracting treatments and outcomes from full-text articles describing clinical trials (entity identification) \emph{and}, ({\it b}) inferring the reported results for the former with respect to the latter (relation extraction). 
We introduce new data for this task, and evaluate models that have recently achieved state-of-the-art results on similar tasks in Natural Language Processing.
We then propose a new method motivated by how trial results are typically presented that outperforms these purely data-driven baselines.
Finally, we run a fielded evaluation of the model with a non-profit seeking to identify existing drugs that might be re-purposed for cancer, showing the potential utility of end-to-end evidence extraction systems.}

\section{Introduction}
\label{section:intro}

Currently, Randomized Controlled Trials (RCTs) pertaining to specific clinical questions are manually identified and synthesized in \emph{systematic reviews} that in turn inform guidelines, health policies, and medical decision-making. Such reviews are critically important, but onerous to produce. Moreover, reliance on these manually compiled syntheses means that even when a systematic review relevant to a particular clinical question or topic exists, it is likely that new evidence will have been published since its compilation, rendering it out of date \cite{bastian2010seventy}.
Language technologies that make the primary literature more \emph{actionable} by surfacing relevant evidence could expedite evidence synthesis \cite{thomas2011applications} and enable health practitioners to inform care using the totality of the available evidence.

Results from individual trials are disseminated as generally unstructured text but will contain descriptions of key components: The enrolled \emph{Population} (e.g., diabetics), the \emph{Intervention} (e.g., beta blockers), the \emph{Comparator} treatment (e.g., placebos), and finally the \emph{Outcomes} measured. Collectively, these are known as the PICO elements. The key findings in an RCT relate these elements by reporting whether a specific intervention yielded a significant difference with respect to an outcome of interest, as compared to a given comparator. 
These results --- reported results for ICO triplets --- are what we aim to extract from trial reports.

Prior work has considered the task of automatically extracting snippets describing PICO elements from articles describing RCTs \cite{jin2018pico,nyeebmnlp,scibert,schmidt2020data,lee2019study}. Other efforts have focused on identifying and analyzing scientific claims \cite{blake2010beyond,kirschner2015linking}. More directly of relevance, prior efforts have also considered the inference problem of extracting the reported finding in an article for a given ICO triplet \cite{deyoung-etal-2020-evidence}.

\begin{figure}
    \centering
    \includegraphics[scale=.45]{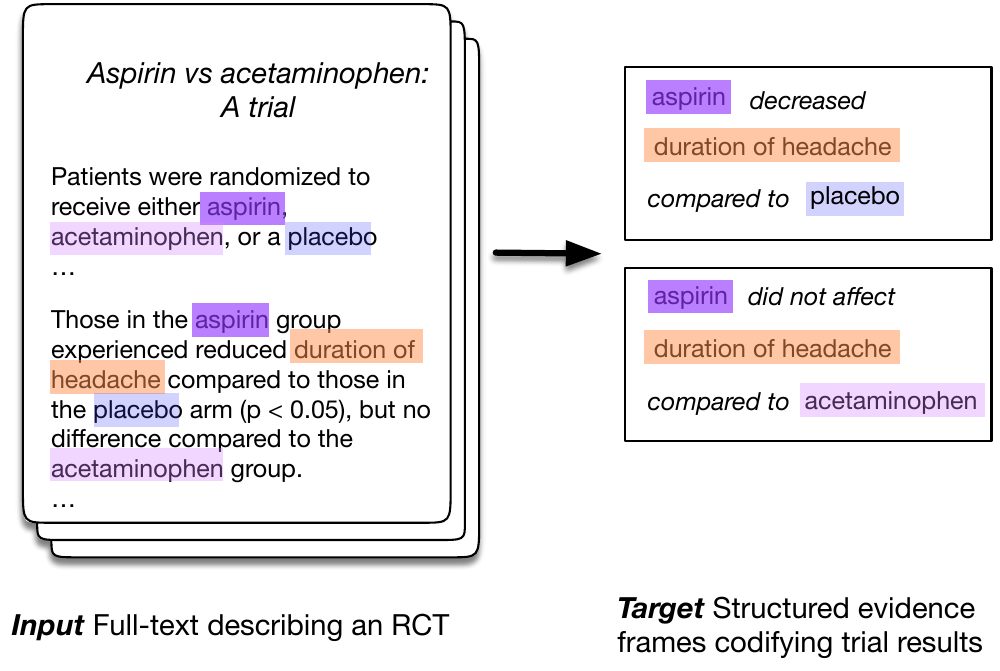}
    \caption{We propose models to extract key clinical entities (interventions, comparators, and outcomes) from reports of randomized controlled trials (RCTs) as well as the reported findings concernig these. Prior work has considered these tasks only in isolation.}
    \label{fig:enchilada}
\end{figure}

Extracting a semantically meaningful structured representation of the evidence presented in journal articles describing results of RCTs is a critical task for enabling a wide range of interactions with the medical literature. 
Patients may, e.g., want to know which side effects are associated with a particular medication; clinicians may wish to know which health outcomes are likely to be affected by a given treatment; and policy makers need to know which healthcare strategies are most efficacious for a particular disease.
However, a usable, end-to-end system must both identify ICO elements within a trial report and infer the findings concerning these: This is the challenge we address in this work.

We design, train, and evaluate systems that extract ICO triplets (specifying which interventions were assessed, and with respect to which comparators and outcomes), \emph{and} infer the corresponding reported findings (Figure \ref{fig:enchilada}), directly from the abstract text. 
This poses difficult technical challenges --- e.g., grouping mentions of the same underlying intervention --- that we aim to address. 
While methods and systems that address the sub-components of this task have been previously proposed, as far as we are award this is the first attempt to design a system for the \emph{end-to-end} evidence extraction task, including ICO identification and inference.

Our contributions in this work are as follows.

\begin{itemize}
    \item We introduce the challenging task of \emph{clinical evidence extraction}, and we provide a distantly supervised training dataset along with a new directly supervised test set for the task.
    \item We propose a novel approach informed by language use, and compare it against current state-of-the-art joint end-to-end NLP models such as DyGIE++ \cite{Wadden2019EntityRA}. 
    \item We evaluate these models both quantitatively and qualitatively. We ablate components and modes of supervision. And we find that while F-scores across models considered appear low in absolute terms (which is unsurprising given the difficulty of the task), domain experts nonetheless find model outputs for an example application --- identifying candidate drugs that might be repurposed for cancer --- useful. 
\end{itemize}

\subsection{End-to-End Evidence Extraction}
\label{section:the-task}

The two core subtasks inherent to evidence extraction are identifying ICO elements and inferring reported relationships between them. 
These tasks may be viewed as instances of Named Entity Recognition (NER) and relation extraction (RE), respectively. 
These general tasks have been extensively studied in the prior work that we build upon here. Traditional approaches to relation extraction use a pipeline approach that entails first extracting relevant entity mentions, and then passing them forward to a relation extraction module. Recent work has proposed performing joint extraction of entities and relations, allowing information to be shared between these related tasks \cite{bmc-joint,DBLP:multihead,DBLP:biaffine}.

When performed jointly with NER, relation extraction is typically treated as a sentence-level task in which interactions are only evaluated between entity mentions that co-occur within a limited range. 
This limitation is crippling in the abstracts of RCT articles; conclusion sentences conveying the relation of interest only explicitly mention the primary intervention 28\% of the time,\footnote{Most trials investigate a particular, often new, intervention of interest and compare this against a placebo or existing standard of care; this is reflected in the framing of the trial. We refer to this intervention as the `primary' intervention, for want of a better term.} often using an indirect coreferent expression such as ``The propofol consumption was similar in the four groups'', or omitting it entirely as in: ``There was no difference in the use of inotropes, vasoconstrictors or vasodilators.''

Further, the directionality of reported results (i.e., whether the intervention \emph{significantly increased}, \emph{significantly decreased}, or induced \emph{no significant difference} relative to the comparator, with respect to an outcome) are conventionally reported with respect to an implicit primary intervention. 
In a statement such as ``The consumption of both propofol and sevoflurane significantly decreased'', the provided evidence for a relation requires knowing which trial arm is the primary intervention, and which the comparator. This information, especially at the abstract level, is typically available only at the beginning of the text. To extract these relations, we must therefore draw upon context derived from the entire document.

Recent work on document-level relation extraction has shown promising results \cite{Wadden2019EntityRA}, but work on medical texts so far has been limited in scope to subareas with comparatively standardized entities (e.g., chemicals or genes) that can be reliably identified and linked to a structured vocabulary via synonym matching \cite{verga-etal-2018-simultaneously}. By contrast, the space of medical interventions is vast, ranging from pharmacological treatments to prescribed animal companions. 

We operationally define \emph{clinical entities} as concepts describing: trial participants (and their conditions); treatments (interventions) that participants were randomized to receive, and; outcomes measured (including measurement scales) to determine treatment efficacy. 
These clinical entities collectively describe the key characteristics of trials and provide the context for interpretation of the reported statistical results.
In evidence extraction we aim to identify $N$-ary relations that capture interactions between treatments and outcomes (Figure \ref{fig:enchilada}).

We say that (\emph{intervention}, \emph{comparator}, \emph{outcome}) triplets exhibit a relation if there is a reported measurement for the outcome with respect to the intervention and comparator. 
We derive candidate relation labels from the Evidence Inference corpus \cite{lehman2019inferring}: The relative effect of the intervention can be \emph{increased}, \emph{decreased}, or \emph{not statistically different}. 
We also consider a relaxed version of this task that considers only binary relations between (\emph{interventions} and \emph{outcomes}. 
This simplification (in which the comparator is implicit) permits direct comparison to existing models that only consider binary relations.
One may interpret this as asking ``what is the comparative effect of this intervention with respect to this outcome, as compared to whatever was used as the baseline".

Each entity is grounded to the abstract as a list of mentions. 
Although entity-focused tasks in related domains often link entities to structured vocabularies such as the Unified Medical Language System (UMLS), such a mapping is difficult in the highly variable setting of general clinical interventions and outcomes. 
Outcomes in particular are often complex combinations of several concepts, such as ``Duration of pain after opening the tourniquet."
For this reason we eschew explicit entity linking for evaluation, and instead say that a predicted relation between extracted mentions constitutes a prediction for the corresponding entities.

To evaluate systems that attempt to perform this task, abstracts must be annotated with: All unique intervention and outcome entities; Mentions (spans) corresponding to each of these, and; The directionality of reported findings for reported comparisons. Independent corpora exist for extracting intervention and outcome spans \cite{nyeebmnlp} and for identifying reported findings concerning these clinical entities \cite{lehman2019inferring}.
However, EBM-NLP does not include groupings of mentions into unique entities, and documents in the Evidence Inference assume ICO entities are \emph{given} as a ``prompt'', and these only sometimes are taken verbatim from the corresponding article.
Furthermore, the Evidence Inference corpus only contains annotations for \emph{some} of the evaluations for which results were reported, i.e., these are non-exhaustive.\footnote{Evidence Inference includes annotations over full-texts, but here we work with an abstract-only subset of this data.} 

Because we do not have direct supervision for this task, we relied on \emph{distant supervision} \cite{mintz2009distant}, i.e., noisy automatically derived `labels'. We derived this from existing corpora. Specifically, we used the EBM-NLP corpus to train a model that we then use to identify (all) entity mentions in the Evidence Inference dataset. For development and testing data, we collected new \emph{exhaustive} annotations from domain experts (medical doctors) on 60 and 100 abstracts, respectively.\footnote{We will publicly release this data alongside publication.}

\section{Methods}

\subsection{Data}
\label{section:data}

As mentioned above, corpora exist for the constituent tasks of extracting interventions and outcomes \cite{nyeebmnlp}, and for inferring results for a given intervention and outcome \cite{lehman2019inferring}, but not for the proposed end-to-end task. 
We therefore use existing datasets and heuristics to derive a relatively large, distantly supervised training set (Section \ref{sec:supervision}). 
We additionally collect relatively small development and train sets explicitly annotated by domain experts (Section \ref{section:eval-annotations}).\footnote{This is expensive, as we acquire \emph{exhaustive} annotations from individuals with medical degrees.}
We will make all data available upon publication

\subsubsection*{Distant Supervision}
\label{sec:supervision}

Documents in the Evidence Inference corpus include triplets comprising: (i) Text describing an intervention; (ii) Text describing a comparator; and (iii) Text describing an outcome. For each such ``ICO'' triplet, a label that indicates the directionality of the reported result is provided, along with a \emph{rationale} span extracted from the source text that provides evidence for this conclusion. For instance, in the illustrative example depicted in Figure \ref{fig:enchilada}, the label for (\emph{aspirin}, \emph{placebo}, \emph{duration of headache}) would be \emph{decreased} and the supporting rationale for this would be the snippet ``Those in the aspirin group experienced reduced duration of headache compared to those in the placebo arm (p$<$0.05)''. 

To identify mentions of the entities, we follow Nye \emph{et al.}\cite{nyeebmnlp} in training a BiLSTM-CRF sequence tagging model on the EBM-NLP data. To encode tokens for the NER model we use SciBERT, which is currently state-of-the-art for this task \cite{scibert}. 
Predicted mentions are then assigned to the provided entity that has the highest cosine similarity with respect to the embeddings induced by SciBERT. Any mention that exceeds a maximum distance threshold is assigned to a new entity that does not instantiate any relations. We tune this distance threshold over the development set.

\subsubsection*{Evaluation Data}
\label{section:eval-annotations} 

\begin{table}[]
\small
    \centering
    \begin{tabular}{lrrrrrr}
              & \multicolumn{2}{c}{Train} & \multicolumn{2}{c}{Dev} & \multicolumn{2}{c}{Test} \\
              \hline
    Abstracts & 1,772  & (1.00) & 60  & (1.00) & 100 & (1.00) \\
    Relations & 4,565  & (2.58) & 200 & (3.33) & 289 & (2.29) \\
    Entities  & 12,556 & (7.09) & 531 & (8.85) & 808 & (8.08) \\
    Mentions  & 29,908 & (16.88) & 1,163 & (19.38) & 1,788 & (17.88) \\
    \end{tabular}
    \caption{Total count (and average per-document) for data types in the distantly supervised train set and expert-labeled dev and test sets.}
    \label{tab:data-stats}
\end{table}

To accurately evaluate performance for this task we collected exhaustive manual annotations over 160 abstracts. For this we hired personnel with medical degrees via Upwork.\footnote{\url{http://www.upwork.com}} 
Annotation was completed in two phases. 
The first step entailed identifying the relations; for this we followed \cite{lehman2019inferring}, except that we requested annotators exhaustively identify \emph{all} relations reported in abstracts (rather than only a subset of them in full-texts).

In the second step annotators marked all intervention and outcome mentions in the abstract, and then grouped these into distinct entities. 
Annotators were instructed to only highlight explicit mentions and to ignore coreferent spans that lacked meaning without the head mention (e.g., ``Group 1"). 
Mentions were grouped only when directly interchangeable in the context of the RCT. 

We hired and trained five expert annotators to perform the second phase, but we only retain annotations from the two most reliable annotators due to the difficulty of the task. 
On a multiply-annotated subset of the development set, inter-annotator agreement for identifying and grouping mentions were calculated using $B^3$, \emph{MUC}, \emph{CEAF}$_{e}$ \cite{pradhan-etal-2014-scoring}. Overall scores were 0.40, 0.46, and 0.42 respectively. 
Each abstract took between 6 and 37 minutes (average 17) to annotate, depending on complexity. The total cost to annotate 160 documents exhaustively was $\sim$\$600.

Across the expert-labeled data the average trial contains 2.62 different treatment arms, necessitating identification of intervention and comparator  entities for each evidence claim rather than at the document level. An additional complication is that sentences presenting a result only make explicit mention of the relevant intervention and comparator arms 37\% of the time, instead using coreferent mentions (e.g. ``Headaches were reduced in the treatment group'') 40\% of the time, or implicit mentions (e.g. ``Overall prevalence of adverse effects was decreased'') 31\% of the time.

\subsection{Modeling}
\label{section:model}

We now describe the models we evaluate for the proposed task: Existing, transformer-based systems that perform extraction and relation extraction jointly and a new approach that we propose informed by how trials tend to report results. 
 
\begin{figure*}
    \centering
    \includegraphics[width=\textwidth]{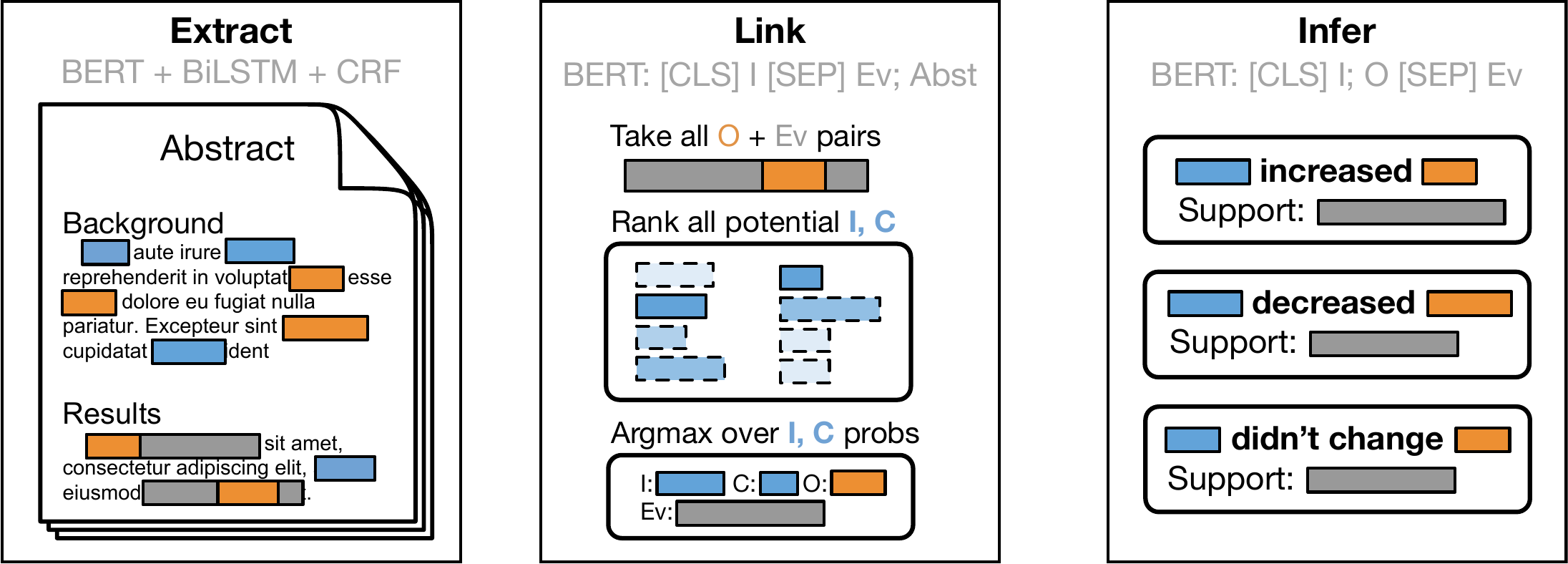}
    \caption{In our proposed Extract, Link, Infer (ELI) method, we first Extract all snippets describing treatments and outcomes, and evidence-bearing sentences. Outcome snippets found within an evidence sentence are then Linked to the most probable abstract-level intervention. The direction of the finding for the selected (intervention, outcome) pair is then Inferred from the evidence-bearing context in which the outcome appeared.}
    \label{fig:schematic}
\end{figure*}
    
\subsubsection*{Joint Models}
\label{section:joint}

Models that perform entity recognition, coreference resolution, and relation extraction jointly may benefit from sharing information across tasks. 
Such models have recently achieved state-of-the-art performance on benchmark tasks, but they can be tricky to train, and it is difficult to incorporate prior knowledge for specific domains into such data-driven models.
For the present task we evaluate two recently proposed models that most closely match our task setting.

The first candidate we consider reports the best known results on the Biocreative V CDR dataset \cite{li2016biocreative}, for which the task is to identify relations between chemicals and diseases in scientific abstracts. 
The Bi-affine Relation Attention Network (BRAN) proposed by \cite{verga-etal-2018-simultaneously} jointly learns to predict entity types and relations across full abstracts, as well as aggregating across the mentions of each entity. This model relies on existing NER labels which are easily obtained for chemicals and diseases, but more challenging in the broader domain of our task.

We also compare to DyGIE++, which recently achieved top results on several scientific extraction datasets, including SciERC, GENIA, and ACE05 \cite{Wadden2019EntityRA,luan-etal-2018-multi}. This model performs joint NER, coreference, and relation extraction, but does so at the sentence level by iterating over all possible mention pairs. This strategy does not readily scale up to processing full abstracts due to the significant increase in the number of mentions. In light of training and data constraints, we disable the coreference module and additional propagation layers in DyGIE++.

\subsubsection*{Our Approach: Extract, Link, Infer}
\label{section:ELI}

Motivated by observations concerning how results tend to be described in trial reports, we propose a new approach that works by first independently identifying (i) spans describing interventions and (ii) snippets that report key results (i.e., that report the observed comparative effectiveness between two or more treatments, with respect to any outcome). 
Trial reports may present findings for multiple intervention comparisons, with respect to potentially many outcomes. 
In a second step we therefore link the identified evidence-bearing snippet to the extracted outcome and intervention to which it most likely pertains.
This Extract, Link, Infer (ELI) approach (Figure \ref{fig:schematic}) therefore effectively works backwards, first identifying evidence statements and then working to identify the clinical entities that participate in these reported findings.

\begin{figure}
    \centering
    \includegraphics[scale=.375]{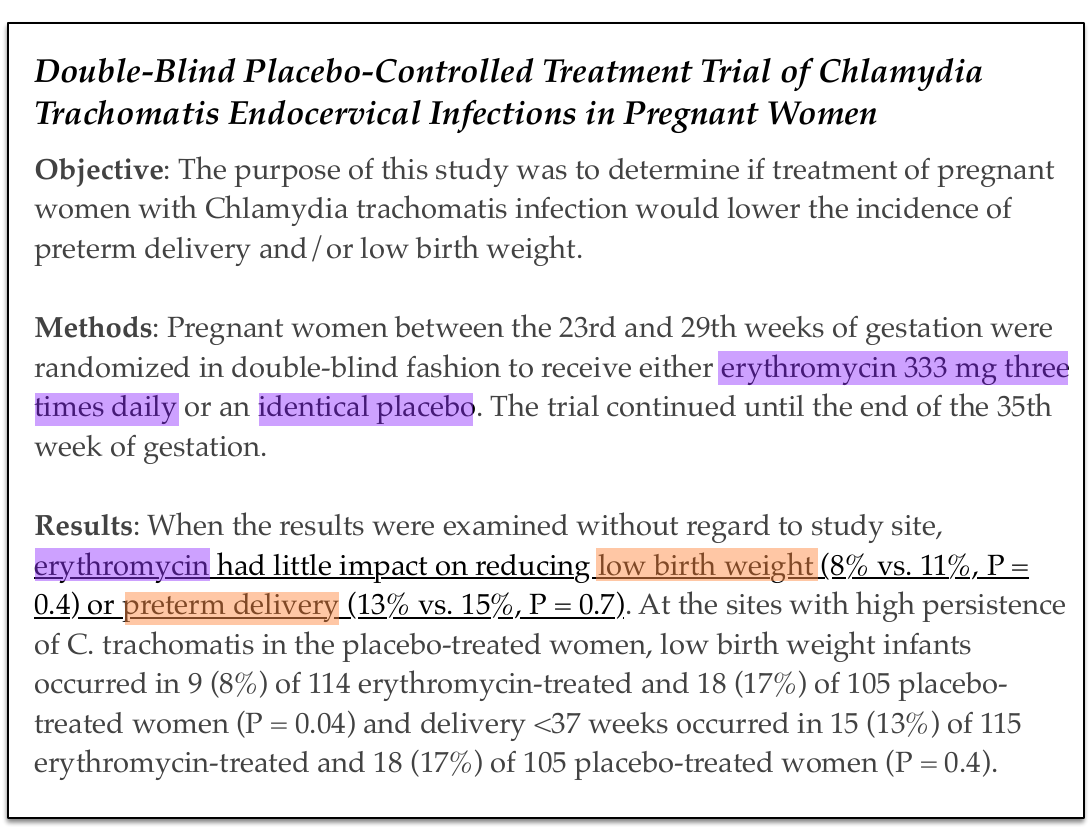}
    \caption{An example abstract. Intervention snippets are highlighted in purple, outcomes in orange. The main evidence-bearing snippet is underlined.}
    \label{fig:example}
\end{figure}

As an illustrative example, consider Figure \ref{fig:example}. 
The main findings are reported in the underlined snippet: ``erythromycin had little impact on reducing low birth weight (8\% vs. 11\%, P = 0.4) or preterm delivery (13\% vs. 15\%, P = 0.7)''. 
There are two outcomes here: ``low birth weight'' and ``preterm delivery''.
We need to link these findings to the primary intervention that they concern.
The two interventions discussed in this abstract are: ``erythromycin 333 mg three times daily'' and ``identical placebo''; the former is the treatment of interest. 
Finally, we can infer the direction of the reported finding for the primary treatment for extracted outcomes: erythromycin yielded \emph{no significant difference} in both outcomes, versus placebo.
  
Following this illustration, in ELI we decompose evidence extraction into independent components that operate in two phases over inputs. Using independent components brings drawbacks: We cannot borrow strength across tasks, and errors will cascade through the system. But it also allows us to explicitly capitalize on domain knowledge about how results are reported in such texts --- as in the preceding example --- in order to simplify training and render the task more tractable. Note that in purely data-driven systems such as BRAN and DyGIE++, there is no natural means to operationalize the intuitive strategy just outlined.

The initial stage of the ELI pipeline comprises two independent tasks. First, all mentions of interventions and outcomes are extracted using the sequence tagging model described in Section~\ref{sec:supervision}. 
Second, all sentences are classified as containing evidence-bearing snippets (or not). 
For this sentence classification model, we add a linear layer on top of SciBERT representations.

We construct training data for this using evidence spans from the Evidence Inference corpus\cite{lehman2019inferring}. 
We take as positive sentences any that overlap any annotated evidence spans; for negative samples we use sentences of similar length from the same document. 
This model realizes 0.97 recall and 0.53 precision on the Evidence Inference test set.\footnote{The modest precision may reflect the fact that evidence snippets are not exhaustively labeled in the dataset.}

These extracted spans are passed forward to a second stage, in which a model attempts to determine which clinical entities are referred to in a particular evidence span. We obviate the need for an explicit model to link outcomes to evidence spans by observing that reported conclusions almost always contain an explicit mention of the relevant outcomes within the same sentence as the stated result. Therefore, we only consider outcome mentions that occur inside one of the extracted evidence spans. In the development set, 87\% of outcome entities are directly mentioned in an evidence statement.

We train a second sentence pair classification model that takes as input an extracted intervention span and an evidence sentence, and predicts if the given intervention is the primary treatment, the comparator, or is unrelated to the given evidence span. We train this model with the (intervention, comparator, evidence span) triplets provided in the Evidence Inference corpus, augmented with synthesized negatives selected to mimic the failure modes of the NER tagger --- extraneous interventions from within the same document, compound phrases involving multiple interventions, and random spans from other locations in the document. This model is then used at test time to select the most probable intervention mention for a given evidence sentence, producing a pair of interacting intervention and outcome mentions linked to the corresponding evidence sentence.

Finally, a linear classification layer is fine-tuned on top of SciBERT that takes the assembled relation candidate and predicts the directionality of the finding with respect to the given intervention and outcome. 
As reported in prior work \cite{lehman2019inferring}, if ground truth evidence spans are given, predicting the direction of the findings reported in these is comparatively easy: Models achieve an F1 score of 0.80 on this three-way classification task.

\subsection{Experimental Details} 
\label{section:exp-details} 

All of our components operate over representations yielded from BERT \cite{devlin-etal-2019-bert}, specifically the pretrained SciBERT \cite{scibert} instance.
We use the Adam optimizer \cite{KingmaB14}, with learning rate $1e^{-3}$. 

For DyGIE++, we use the default configuration, except: We increase the loss weight for relations from 1 to 10, we disable the coreference module, and we disable relation propogation. DyGIE++ fine-tunes BERT parameters, using the BertAdam optimizer \cite{devlin-etal-2019-bert}. We truncate inputs to 300 tokens (roughly the mean input length), discarding mentions beyond this.


\section{Results}

We present quantitative results for the proposed end-to-end task in Section \ref{section:main-results}, comparing ELI to modern transformer-based joint models that represent the SOTA for similar tasks. We then ablate the components of ELI in Section \ref{section:ablations} to characterize where the system does well and where it fails, highlighting directions for improvement. 

All systems we evaluate perform relatively poorly in absolute terms on this challenging task. To better characterize system performance, we therefore enlist domain experts from a non-profit organization to qualitatively assess the accuracy and potential utility of model outputs. 

\subsection{Main Task Results} 
\label{section:main-results} 

\begin{table}[]
\small
    \centering
    \begin{minipage}{.45\linewidth}
    \centering
    \begin{tabular}{llll}
         \textbf{Entity Extraction}   &   P &   R &  F1 \\
               \hline
    DyGIE++     & 0.45 & 0.47 & 0.46 \\
    ELI         & 0.46 & 0.69 & 0.55 \\
    \\
    \end{tabular}
    \end{minipage}
    \begin{minipage}{.45\linewidth}
    \centering
    \begin{tabular}{llll}
         \textbf{Relation Inference}   &   P &   R &  F1 \\
               \hline
    BRAN       & 0.05 & 0.41 & 0.08 \\
    DyGIE++    & 0.24 & 0.13 & 0.17 \\
    ELI        & 0.33 & 0.31 & 0.32  
    \end{tabular}
    \end{minipage}
    \caption{System performances on the intermediate entity extraction and the end-to-end inference tasks.}
    \label{tab:relation-extraction-results}
\end{table}

We report results for both the intermediate aim of entity extraction and the final task of inferring relations between entities in Table \ref{tab:relation-extraction-results}. As expected, the task of mapping raw inputs to structured results relating clinical entities is quite challenging, yielding low absolute numbers for current state-of-the-art joint models. While DyGIE++ achieves moderate success (0.46 F1) with respect to identifying mentions of the relevant clinical entities, performance with respect to the final task of identifying the relations between them leaves much to be desired.

Our proposed method, ELI, fares considerably better than DYGIE++, with relative F1 increases of 20\% and 88\% for entity extraction and relation inference respectively.
That said, in absolute terms the performance of ELI is also low (0.32 F1), highlighting the challenging nature of the task. However, we note that we note that even while performance seems poor here in absolute terms with respect to F1, our evaluation in which domain experts directly assess model outputs (Section \ref{section:cwr4c}) suggests that the ELI system is already practically useful for downstream applications.

One of the primary potential drawbacks of employing a series of independent models is that errors made in an early stage will propagate through the system, resulting in lower quality inputs and higher error rates for the subsequent models. In the next Section we ablate components of ELI to highlight potential means of improving this performance.

\subsection{Ablations and Analysis} 
\label{section:ablations}

%

To investigate the sources of error that accumulate over the course of the ELI modeling pipeline, we evaluate each component model individually.

The extraction phase, which consumes raw unlabeled documents, produces per-token labels for intervention and outcome mentions. We achieve 0.55 F1 (0.69 recall) at the token level. However, a system does not need to extract every mention of an entity in order to arrive at the correct conclusions. As long as at least one of an entity's mentions are correctly extracted, it is still a candidate for participating in different relations. We therefore evaluate clinical extraction at the entity level as well, where an entity is marked as extracted as long as we identify at least one of its mentions. While this is more permissive from a recall perspective, we treat any extracted span that is not a mention of a ground-truth entity as a false positive. This produces a pessimistic view of the true precision, since the metric penalizes repeated extraction of spans that represent a single false entity. At the entity level we improve extraction scores to 0.69 F1, and more importantly recall increases to 0.85.

Extraction of evidence sentences is very high recall (0.98) with middling precision, and we rely on the linking phase to correctly reject evidence sentences that do not contain conclusive statements about specific outcomes.

We next isolate the performance of the linking phase by providing ground-truth evidence sentences and entity mentions as inputs. The model ranks all intervention and comparator entities, linking the most probable for each evidence sentence. We observe that 78\% of outcome entities are contained within an evidence sentence (and therefore automatically linked), and the model reaches 75\% and 70\% accuracy for selecting the correct intervention and comparator, respectively.

The inference model is then given all ground-truth outcome spans and their corresponding evidence sentences. Instances for which \emph{no significant difference} is reported are easy to classify (0.93 F1), while significant changes are more difficult (0.71 F1). The majority of mistakes come from mislabeling decreases as increases; in many cases this error is caused by cases in which a decrease in an undesirable outcome is reported positively (e.g. ``adverse effects were improved in the treatment group'').

\begin{table}[]
    \centering
    \begin{minipage}{.33\linewidth}
        \centering
        \begin{tabular}{l lll}
            \textbf{Extraction}       &   P &   R &  F1 \\
                   \hline
            tokens     & 0.45 & 0.69 & 0.55 \\
            entities   & 0.58 & 0.85 & 0.69 \\
            evidence   & 0.45 & 0.98 & 0.62 \\
        \end{tabular}
    \end{minipage}
    \begin{minipage}{.25\linewidth}
        \centering
        \begin{tabular}{l l}
            \textbf{Linking} & Acc \\
                   \hline
            interventions & 0.75 \\
            comparators   & 0.70 \\ 
            outcomes      & 0.78 \\
        \end{tabular}
    \end{minipage}
    \begin{minipage}{.33\linewidth}
        \centering
        \begin{tabular}{l lll}
            \textbf{Inference} &   P &   R &  F1 \\
                   \hline
             increased      & 0.64 & 0.90 & 0.75 \\
             decreased      & 0.85 & 0.54 & 0.66 \\
             no difference  & 0.93 & 0.94 & 0.93 \\
        \end{tabular}
    \end{minipage}
    \caption{Performance of each ELI module on the test set, if given ground-truth inputs.}
    \label{tab:eli-ablation}
\end{table}

\subsection{Application Example: Helping RebootRX Find (Potential) Cancer Treatments}
\label{section:cwr4c} 

To assess the practical utility of the presented task and systems, a domain expert at \url{rebootrx.org}, a non-profit organization that seeks to identify previously studied drugs from the literature that might be repurposed to treat cancer, evaluated outputs from the pipeline system on 20 abstracts from RCTs investigating cancer treatments. 

The clinical entities in this sub-domain are particularly challenging to extract and differentiate: Interventions typically consist of compound treatments with complicated dosage schedules, and similar outcomes are often measured repeatedly at different times. RebootRX seeks to identify all RCTs in which specific interventions were used to evaluate outcomes of interest; this is an information retrieval problem that maps directly on to our proposed task of evidence extraction.

The domain expert was asked to assess the per-document recall and precision of the extracted relations on a five point Likert scale, and to note which aspects of erroneous predictions were incorrect. Overall, the annotator scored the ELI system at 80.0\% recall and 63.8\% precision. They noted that a mistake in predicting the directionality of the results accounted for 47.6\% of the extraction errors. Therefore, despite the ostensibly low absolute scores on the strict evaluation performed above using our test set, these results indicate that even current models for the proposed task are useful in a meaningful, real-world setting.

\section{Discussion}

Automating structured evidence extraction has the potential to provide better access to emerging clinical evidence. In the immediate future, this may reduce the manual effort needed to produce and maintain systematic reviews. 
Thinking longer term, if we can improve the accuracy of such end-to-end systems we may be able to eventually realize \emph{living} information systems, where health professionals could draw upon real-time assessments of the evidence to inform their decision-making.

Realizing this potential will require improving methods for automated evidence extraction, which in turn necessitates 
addressing NLP challenges relating to joint extraction and grouping of heterogeneous treatment and outcome mentions from relatively lengthy inputs, and inference concerning the reported relations between these. This domain and setting poses several interesting obstacles such as frequent use of coreference and implied mentions, and relations requiring document-level context.

 A system for extracting rich, structured data from a completely unlabeled document has many opportunities to make mistakes, and this composition of complex tasks provides significant challenges for current state of the art systems. Our proposed model achieves strong results on each component task, and even so overall performance leaves room for improvement. Our hope is that this difficult, important task motivates innovations to meet these challenges.

One avenue of future research that this motivates is how one might integrate intuitions about language use in particular tasks into end-to-end, joint systems. Other promising leads include extending ELI to incorporate distantly supervised labels, as well as refinement of the supervision strategy.

\section{Conclusion}

We have proposed the end-to-end task of automatically extracting structured evidence --- interventions, outcomes, and comparative results --- from trial reports. 
This differs from prior work which has considered the tasks of data extraction and inferring results separately. 

We introduced new data for this task (which we will make publicly available), including a large distantly supervised train set and expert annotated development and test sets.
The latter feature \emph{exhaustive} annotations, inclusive of all ICO triplets for which results have been reported in the corresponding abstract.
Using this data, we evaluated state-of-the-art joint NLP models for this challenging task and proposed a new modular method --- Extract, Link, Infer (ELI) --- motivated by observations about how authors convey findings in trial reports.
This model yielded superior performance on the task, presumably due to its implicit incorporation of domain knowledge concerning how trial reports are structured.

We performed a fielded evaluation in collaboration with a non-profit (RebootRX) that is interested in identifying results from previously conducted studies on drugs that might be used to treat cancer.
This small study suggested that despite low absolute performance metrics, the proposed model can be useful in practice.

\section*{Acknowledgements}

This work was supported in part by National Science Foundation (NSF), grant 1750978; and by the National Institutes of Health (NIH), grant R01-LM012086.

\bibliographystyle{unsrt}

\end{document}